%% file: acl_submission.tex
\documentclass{article} 
\usepackage[preprint]{acl}
\usepackage{times}

\usepackage{microtype}
\usepackage{hyperref}
\usepackage{url}
\usepackage{booktabs}
\usepackage{xspace}
\usepackage{lineno}
\usepackage{listings}
\usepackage{booktabs}
\usepackage{multirow}
\usepackage{svg}
\usepackage{framed}
\usepackage{tablefootnote}

\definecolor{darkblue}{rgb}{0, 0, 0.5}
\hypersetup{colorlinks=true, citecolor=darkblue, linkcolor=darkblue, urlcolor=darkblue}

\usepackage{enumerate}
\usepackage{graphicx}

\usepackage{amsthm}
\usepackage{amsmath}
\usepackage{amsfonts}
\usepackage{xpatch}
\makeatletter
\xpatchcmd{\@thm}{\thm@headpunct{.}}{\thm@headpunct{}}{}{}
\makeatother
\newtheorem*{lemma}{lemma}
\usepackage{stmaryrd}

\title{Lemmanaid: Neuro-Symbolic Lemma Conjecturing}

\author{Yousef Alhessi\textsuperscript{1}, Emily First \& Sorin Lerner \\
Department of Computer Science and Engineering\\
University of California, San Diego, USA\\
\texttt{\{yalhessi,emfirst, lerner\}@ucsd.edu} \\
\AND
S\'{o}lr\'{u}n Halla Einarsd\'{o}ttir\textsuperscript{1}
, George Granberry, Moa Johansson \& Nicholas Smallbone \\
Department of Computer Science and Engineering\\
Chalmers University of Technology \& University of Gothenburg\\
Gothenburg, Sweden\\
\texttt{\{slrn,georgegr,moa.johansson,nicsma\}@chalmers.se} \\
\textsuperscript{\textbf{1}}Joint first authorship
}

\newcommand{\tool}{\textsc{Lemmanaid}\xspace}

\begin{document}

\maketitle
\input{main}

\section*{Limitations}
Due to limitations in computational resources, we have evaluated the Lemmanaid framework with smaller LLM backends, ranging between 1B to 6.7B, for two different model families, DeepSeek and Llama 3. Each instance of Lemmanaid requires fine-tuning both the template-generation model, and a direct neural conjecturing model. This in turn, was done for two (or three) different prompting strategies (see e.g. table \ref{fig:rq1-main}), including either both type and function definitions, or only one of them. Further prompting experiments, and wider beam searches were beyond the scope of our computational budget. Furthermore, for the largest model (Deepseek-coder-6.7b) we suspect additional computational resources might have revealed slightly better hyper-parameter settings, which with additional fine-tuning could have improved those results further.

We have evaluated \tool by comparing it to gold-standard human written lemmas from Isabelle formalizations, motivated by the assumption that human domain experts have a good idea of what is interesting and useful in each different formalization. Evaluating whether the conjectures are actually useful for other, yet unknown, downstream tasks was not the goal of the evaluation, nor the question we aimed to answer in this paper. Assessing the usefulness of a tool like \tool for human users working on real formalization tasks would indeed be a very interesting follow-up study warranting a paper in its own right, but is beyond the scope of the current paper.

As described in \S\ref{sec:qualitative}, proofs of the conjectures generated by \tool have only been attempted by the fully automated tactic Sledgehammer. While powerful, any conjectures requiring a more specialized tactic-based proof remain open. In this paper, our focus was a comparison across a wide range of formalizations against their gold-standard lemmas, and we leave deeper analysis and interactive proof attempts of open conjectures in each theory as future work. 

We conducted a smaller evaluation of using commercial high-powered LLMs as backends for \tool as well as for direct conjecturing. We limited this evaluation to a smaller set for two reasons: First, we wanted to at least minimize the amount of test-data leakage by choosing formalizations entered into the Isabelle AFP after the training cut-off dates of the LLMs. Secondly, we needed to keep to a limited budget. While we attempt to conduct the comparative experiments under as similar conditions as possible, we once again emphasize that there are inherent differences in the models considering computational resources and cost, training, and input prompt token usage. 




\bibliography{refs}

\appendix
\input{Appendices/template_language}
\input{Appendices/resources}



\input{Appendices/greedy_results}

\section{Breakdown of Results per Theory}
\label{app:breakdown-theories}
\input{Appendices/HOL_results}
\input{Appendices/AFP_results}

\input{Appendices/lemma_statistics}

\input{Appendices/qualitative_results}

\input{Appendices/commercial}
\input{Appendices/qs_octonions}
\end{document}

%% file: main.tex
\input{main_contents/0_abstract}
\input{main_contents/1_intro}
\input{main_contents/2_related}
\input{main_contents/3_approach}
\input{main_contents/4_eval}
\input{main_contents/5_conclusion}


%% file: main_contents/0_abstract.tex
\begin{abstract}
\vspace{5ex}
Mathematicians and computer scientists are increasingly leveraging \emph{proof assistants} to formalize and check complex proofs, a task that demands substantial expertise. Can we lower the bar by automating the conjecturing of helpful, interesting and novel lemmas?
We present the first neuro-symbolic lemma conjecturing tool, \tool, designed to discover conjectures by drawing analogies between mathematical theories.
\tool uses a fine-tuned LLM to generate \emph{lemma templates} that describe the shape of a lemma, and symbolic methods to fill in the details. We compare \tool against the same LLM fine-tuned to generate lemmas directly, as well as a fully symbolic conjecturing method.
On test sets from Isabelle's HOL library and Archive of Formal Proofs (AFP), \tool consistently outperforms both neural and symbolic methods. Using DeepSeek-coder-6.7B as a backend, \tool discovers 50\% (HOL) and 29\% (AFP) of the gold standard lemmas, increasing to 55\% and 35\% when ensembling prompting strategies.
In a case study on Octonions, \tool discovers 79\% of the gold standard lemmas, compared to 62\% for neural-only and 23\% for the state of the art symbolic tool.
Furthermore, in a targeted comparison, \tool  discovers more gold standard lemmas than both Claude Opus 4.5 and GPT-5.2.
Our results show that \tool can conjecture a significant number of interesting lemmas across complex formalizations in mathematics and computer science.

\end{abstract}

%% file: main_contents/1_intro.tex
\section{Introduction}

Proof assistants like Lean~\cite{lean} and Isabelle~\cite{isabelle} are becoming increasingly important among research mathematicians who want intricate proofs verified for guaranteed correctness. Following the formalization of Kepler's conjecture \cite{keplerformalization}, there have been several projects formalizing recent results in research mathematics \cite{scholtze,liquidtensor, gowers2023conjecturemarton, tao}. 
However, formalizing mathematics in a proof assistant is a non-trivial task. This in turn has sparked research in areas where large language models might be of assistance, primarily in autoformalization \cite{wang2018neuraltranslation,szegedy2020autoformalisation,wu2022autoformalization} and formal proof synthesis \cite{polu2020generativelanguagemodelingautomated, jiang2022thor,jiang2023draft,first2023baldur}. In these settings, the LLM and proof assistant complement each other: LLMs can provide a more flexible and powerful proof search while hallucinations in proofs are caught by the proof assistant and easily discarded.

\begin{figure}[tbp]
\begin{center}
\vspace{10ex} 
\includegraphics[width=0.98 
\columnwidth]{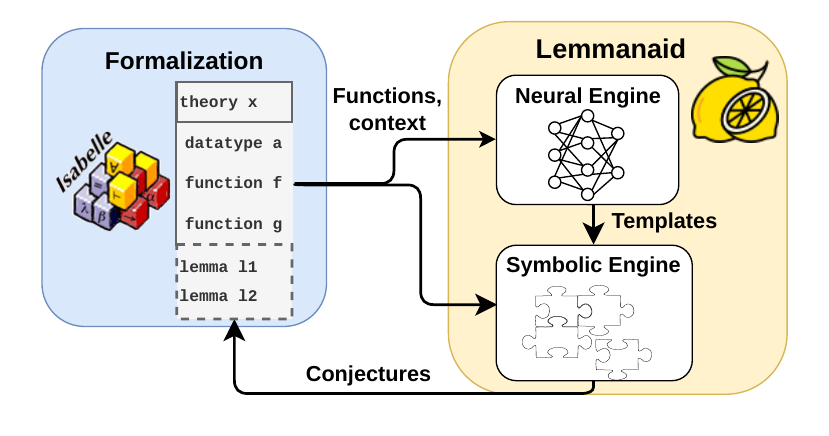}
\end{center}
\vspace{-3ex} 
\caption{High-level overview of \tool.}
\vspace{-3ex} 
\label{fig:vision}
\end{figure}

We address a different challenge, \emph{conjecturing}: learning to construct new, interesting, and useful lemmas. This is an important and long standing challenge in AI for mathematical reasoning \cite{amsystem, graffiti, productiveuse, hrsystem}, yet it remains an underexplored area \citep{yang2025position}. Automatically discovered lemmas can both aid a human user working on a mathematical formalization and strengthen automated theorem provers.
General conjecturing must work across a broad range of theories, spanning computer science, logic and mathematics.
However, standard benchmarks for formal mathematical reasoning have been derived from high school and undergraduate level mathematics competitions and standard textbooks, like miniF2F \cite{zheng2022miniff}, PutnamBench \cite{tsoukalas2024putnambench} and ProofNet \cite{azerbayev2023proofnet}. We move beyond these limited domains by leveraging real formalizations by researchers in both mathematics and computer science.

In this work we introduce \tool: a novel neuro-symbolic approach and tool for general lemma conjecturing across a broad range of mathematical theories.
\tool combines LLMs' generalization capabilities with symbolic tools to generate lemmas by a process of \emph{analogy}.
The LLM suggests generic \emph{templates} describing learned families of analogous lemmas, which are then instantiated to concrete theorems symbolically (Figure \ref{fig:vision}).
Previous work has shown that such families of analogous lemmas indeed occur in proof
assistant
libraries~\cite{lolcicm,proofpatternrec}.

Existing conjecturing approaches rely on either a neural or a symbolic engine.
Neural methods may generate lemmas that are repetitive, redundant or trivial, or hallucinate symbols that are undefined in the formalization \citep{urban2020,rabe2021,explconjllm}. Scaling the model is a commonly attempted remedy, albeit computationally expensive and most effective in domains where background libraries have already been formalized, such as the theories needed for math olympiad problems~\cite{zheng2022miniff}.
Symbolic methods, on the other hand, can be designed and programmed to avoid repetition and redundancy.
Prior symbolic tools ~\citep{quickspec,roughspec,ThyExpDedSyn} have been used to successfully discover lemmas needed in automated (co-)inductive provers~\citep{hipster,cohipster,vampspec,cclemma} over (co-)recursive datatypes. However, symbolic tools are limited in the shape, size and domain of lemmas they can generate, and scale poorly to large inputs sets.
\tool addresses these shortcomings in its novel neuro-symbolic lemma conjecturing approach.
The LLM suggests appropriate analogous lemma-patterns likely to be relevant for the theory at hand. The symbolic engine ensures correctness and novelty, while keeping the search space manageable.
In this way, we leverage the best of both neural and symbolic methods.
As far as we are aware, this is the first work focusing on neuro-symbolic lemma conjecturing.

\tool is a bottom-up conjecturing tool. Automated conjecturing broadly follows two main methodologies~\cite{LemmaDiscoverySurvey}: \emph{top-down} and \emph{bottom-up}.
Top-down methods find relevant lemmas for a specific proof attempt~\cite{productiveuse, proofpatternrec}, with recent work in test-time reinforcement learning producing custom curricula to train specialized neural provers~\cite{dong2025stpselfplayllmtheorem, alphaproof, chen2025seedproverdeepbroadreasoning}.
Success is measured by increased proof success rate (of original proofs).
In contrast, bottom-up methods like \tool find interesting lemmas from a set of definitions, without any predefined top-level goal~\cite{amsystem, mathsaid, urban2020, explconjllm}.
Evaluating bottom-up methods requires defining a notion of conjecture \emph{interestingness}, typically using human-written formalizations as a gold-standard~\cite{isascheme,quickspec,cohipster,urban2020}.
Thus, we evaluate \tool by its coverage of unseen test-set lemmas within a subset of Isabelle's HOL library\footnote{\url{https://isabelle.in.tum.de/dist/library/HOL/index.html}} and its Archive of Formal Proofs (AFP)\footnote{\url{https://www.isa-afp.org}}.
\paragraph{Contributions}
We show that the \tool approach compares favorably to the results achievable using purely neural or purely symbolic conjecturing.
The main contributions of our work are:
\begin{itemize}
    \item \tool, the first neuro-symbolic lemma conjecturing approach that uses an LLM to suggest templates and a symbolic engine to instantiate templates as candidate lemmas.
    \item A broad evaluation of \tool on the Isabelle proof assistant's HOL and AFP libraries. This goes beyond evaluations of prior tools that have focused on specific mathematical domains.
    \item A comparison to an existing symbolic method, QuickSpec, and to neural LLM-based lemma conjecturing models we create, showing that \tool outperforms these methods and is complementary.
    \item A comparison to instruction-tuned commercial LLMs on very recent entries in the AFP, showing that \tool outperforms these models with smaller local models and token usage.
\end{itemize}

%% file: main_contents/2_related.tex
\section{Related Work}
\paragraph{Proof Assistants and Autoformalization}
Proof Assistants~\citep{isabelle, lean} can check proofs in mathematics and computer science for correctness. A user must \emph{formalize} their theory by translating it into the format language of the proof assistant and then interact with the system to construct a proof with calls to \emph{tactics}, each executing and checking a part of the proof. Tools like Sledgehammer~\citep{sledgehammer} can automate many simpler proofs. 

Formalization remains difficult, driving research in \emph{autoformalization}: translating definitions, theorems and lemmas written in natural language to the formal language of a proof assistant~\citep{wang2018neuraltranslation, szegedy2020autoformalisation, wu2022autoformalization}. 
Autoformalization on realistic definitions is still a challenge for LLMs~\citep{zhang-etal-2025-autoformalization}.
\tool serves as an excellent complement: autoformalized definitions can be passed to \tool to generate a richer initial formalization. 

\paragraph{Proof Synthesis}
Even with definitions and statements formalized, constructing the required proofs from various tactics is non-trivial, even with the help of tools like Sledgehammer. This has motivated work on various LLM-driven methods for synthesizing proof scripts for proof assistants such as Isabelle or Lean, either step by step, whole proofs at once, or via proof sketches \citep{polu2020generativelanguagemodelingautomated, wang2024legoprover, ren2025deepseekproverv2advancingformalmathematical,alphaproof,chen2025seedproverdeepbroadreasoning, wang2025kiminaproverpreviewlargeformal,lin2025goedelproverfrontiermodelopensource, jiang2022thor,first2023baldur,jiang2023draft}. 
Retrieval augmentation has been used to complement tactic search by fetching suitable lemmas and similar proofs from existing libraries \citep{leandojo,thakur2024an, 10.1109/ICSE55347.2025.00161}, with  \citet{kumarappan2025leanagent} also targeting continuous learning on growing libraries. 
While our work does not focus on producing proofs, we see it as an excellent complement for future integration with systems like the above. 

\paragraph{Neural Conjecturing and Reinforcement Learning}
Recent top-down conjecturing methods have successfully been used in \emph{test time reinforcement learning} as a means to generate additional data with the aim of creating a curriculum for training a neural prover \cite{dong2025stpselfplayllmtheorem, alphaproof,chen2025seedproverdeepbroadreasoning}, which has contributed to gold medal performances on International Math Olympiad problems \cite{alphaproof, chen2025seedproverdeepbroadreasoning}. Here, conjectures are typically variants of some given \emph{seed statement} of interest. 
\tool, on the other hand, uses a bottom-up conjecturing approach and targets a different use-case. It aims to make novel suggestions directly from \emph{definitions}, interesting to a human doing a formalization, rather than generating variants of a given statement.

Other methods combining neural conjecturing and reinforcement learning have demonstrated success in specific domains
, such as inductive proofs about synthesized programs that generate integer sequences
~\citep{gauthier2025learningconjecturingscratch}. 
LEGO-prover attempts to introduce intermediate proof statements as reusable lemmas for other problems in the same domain~\cite{wang2024legoprover}. Minimo treats conjecturing as a reinforcement learning game in simple propositional logic, arithmetic, and group theory, with well-formed conjectures generated neurally via constrained decoding~\cite{poesia2024learning}. Fermat combines symbolic rules and reinforcement learning to form conjectures in elementary number theory and on finite fields~\cite{tsoukalas2025learning}, and is evaluated against a gold-standard set of conjectures while simultaneously developing an interestingness function over conjectures.
In contrast, \tool targets the broad range of theories from research in computer science and mathematics represented in proof assistant libraries, and aims to avoid domain specific learning by generating generic templates as an intermediate step. 

\paragraph{Templates for Synthesizing Conjectures}
Following the observation that many mathematical theories share analogical lemmas of similar shapes,  \citet{theorema} first proposed to use \emph{templates} as human-provided guidance in mathematical theory exploration, by conjecturing interesting lemmas by analogy to known shapes. This has been implemented in a range of symbolic lemma conjecturing systems, including some targeting  Isabelle/HOL 
\citep{isascheme, proofpatternrec, mathsaid, roughspec, NagashimaTemplateBased}. A similar technique called \emph{sketching} has also been applied in the domain of program synthesis \citep{sketching2019}. In the above works, templates have typically been provided by the human user. \citet{nye2019learninginferprogramsketches} used a neural network to propose program sketches which were then filled in by a symbolic program synthesizer, thus sharing some features with our work.

%% file: main_contents/3_approach.tex
\section{The \tool Approach}

\begin{figure*}[t]
\centering
\includegraphics[width= \textwidth]{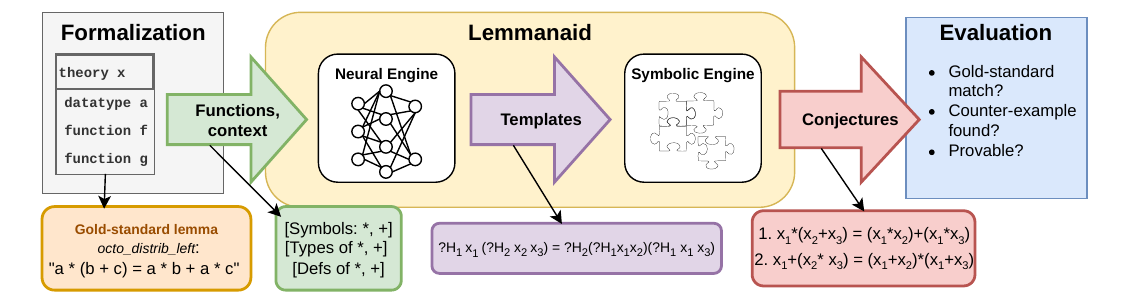}
\caption{Overview showing how \tool conjectures lemmas and how we evaluated generated conjectures.}
\vspace{-2ex} 
\label{fig:overview}
\end{figure*}

 We have implemented a tool, \tool, for template-based conjecturing in Isabelle/HOL.
We envision a user working on a new mathematical formalization; having defined some functions, types and other concepts, and perhaps a few theorems about these.
This collection of definitions, which we refer to as a (partial)
\emph{theory}, serves as input to the conjecturing system.
\tool then outputs conjectures that are likely to be useful in this context, allowing the user to make progress in their formalization, or better understand the behavior of the theory they have defined so far. This happens in two stages: First (neural part), the partial theory is given as input to an LLM which outputs templates likely to be relevant. Second (symbolic part), \tool searches over possible instantiations of those templates in the current theory, to produce concrete conjectures.
See Figure~\ref{fig:overview} for an overview.

We make all code, experimental scripts, data, and models publicly available online\footnote{\url{https://anonymous.4open.science/r/lemmanaid-emnlp}}. See Appendix \ref{app:computing} for details about computing resources and replication parameters used in our evaluation.

\subsection{\tool's Symbolic Engine}\label{sec:symbolic-engine}

\paragraph{Template Language}
 A \emph{template} is an abstraction of a mathematical statement with most of the concrete operators replaced by schematic variables called \emph{holes}. The template captures the overall structure of a statement, with the intention that a template represents a family of analogous lemmas. Basic logical operators, such as equality and quantifiers, are part of the template to retain this high-level structure \footnote{See Appendix \ref{app:templates} for details about the standard logic symbols included in our template language.}. We simply implement templates using the standard internal \texttt{term} type in Isabelle/HOL \cite{Isabelle-implementation}.
  As an example, consider the following lemmas about the Octonionic product and sum functions from the Isabelle ATP formalization about the octonions, an eight-dimensional extension of complex numbers \citep{Octonions-AFP}:

\begin{lemma}
octo\_product\_noncommutative: \\ 
\(\neg(\forall x\; y :: octo. (x * y = y * x))\)
\end{lemma}
\begin{lemma}
octo\_distrib\_left: \\ 
\(a * (b + c) = a * b + a * c\; \texttt{for}\; a\; b\; c ::octo\)
\end{lemma}
\begin{lemma}
octo\_assoc\_plus: \\ 
\(a + (b + c) = (a + b) + c\; \texttt{for}\; a\; b\; c ::octo\)
\end{lemma}
%
If we abstract away the function symbols operating on octonions, and rename the variables according to our template abstraction method, we obtain the three templates:
\begin{small}
\begin{equation}
\label{eq:template1}
   \neg(\forall\ y_0\ y_1.\ ?H_1\ y_0\ y_1 =\ ?H_1\ y_1\ y_0)
\end{equation}
\begin{equation}
   \label{eq:template2}
?H_1\ x_1 \ (?H_2\ x_2\ x_3) =\\ ?H_2\ (?H_1\ x_1\ x_2)\ (?H_1\ x_1\ x_3)
\end{equation}
\begin{equation}
\label{eq:template3}
?H_1\ x_1 \ (?H_1\ x_2\ x_3) =\ ?H_1\ (?H_1\ x_1\ x_2)\ x_3
\end{equation}
\end{small}
Note how the function symbols $*$ and $+$ operating on octonions have been replaced by the holes $?H_1$ and $?H_2$, and the variables $a,b,c,x,y$ standardized to $y_0,y_1,x_1,x_2,x_3$.
Note that logical symbols, such as negation $\neg$, universal quantifier $\forall$ and equals sign $=$ remain in the template structure.
Technically, templates are implemented as instances of Isabelle/HOL's term datatype\footnote{See section 2.2 of the Isabelle/Isar Implementation Manual \url{https://isabelle.in.tum.de/dist/Isabelle2025/doc/implementation.pdf}}.

Our template language represents analogous lemma statements, that have been abstracted and normalized in the following way:
\emph{Function symbols} are replaced with a hole, represented as $?H_k$ where $k$ is a positive integer label. Note that the type of the hole is also an abstraction of the type of the original symbol, with concrete types replaced by type variables. In our examples, the holes in the templates above have types that match any function with two arguments. Each occurrence of a particular function symbol is replaced by the same hole label, even in the case of polymorphic functions where different occurrences of the symbol may have different types. \emph{Variables} are renamed to $x_k$ (or $y_k$ in the case of bound variables), where $k$ is a non-negative integer label, and their types set to match those of the corresponding holes.

\paragraph{Template Instantiation}
We can \emph{instantiate} or \emph{fill} the holes in a template by any operator with a matching type. For instance, template \ref{eq:template1} can be instantiated using \emph{any} binary operator, such as subtraction on real numbers or concatenation on lists, to obtain a new conjecture. If the operator indeed is non-commutative, we obtain a correct lemma analogous to the lemma \textit{octo\_product\_noncommutative}. In this way, templates are generalizations of lemma statements: each template matches many different lemmas that may apply in different theories.

\tool's symbolic engine takes as input a template and a list of function names. It then searches over the possible instantiations of the holes in the template using the given functions (note that unlike holes the variables are fixed at instantiation, and not replaced by any further terms). Each indexed hole is replaced by one of the given function symbols while ensuring that the resulting candidate lemma is well typed. This may lead to a number of different candidate lemmas being produced.
This instantiation task can be viewed as a constraint satisfaction problem (CSP): holes are variables whose domains are function symbols, while type signatures of symbols and arities of holes impose constraints on valid instantiations.
We use standard forward checking \cite{forwardchecking} to efficiently prune the search space.


Revisiting our example, consider again template \ref{eq:template3} (associativity).
Suppose we give this template to \tool's symbolic engine, along with the functions $+$ (addition), $-$ (subtraction), $sin$, $cos$ and $\;\hat{}\;$ (exponentiation) for real numbers, and the functions $len$ (the length of a list), $rev$ (reversing a list) and $@$ (concatenate two lists).
In the hole-filling step, it will come up with the following candidates:
\begin{align*}
    (x_1 + x_2) + x_3 &= x_1 + (x_2 + x_3)  \\ 
    (x_1\ @\ x_2)\ @\ x_3 &= x_1\ @\ (x_2\ @\ x_3) \\
        (x_1 - x_2) - x_3 &= x_1 - (x_2 - x_3) \\ 
    (x_1 ^ {x_2}) ^ {x_3} &= x_1 ^ {x_2 ^ {x_3}}
\end{align*}
Note how $sin$, $cos$, $len$, and $rev$ cannot be used to fill the hole $?H_1$, since $?H_1$ is applied to two arguments and therefore requires a binary operator. Also note that only the instantiations with $+$ and  $@$ result in valid lemmas, while the other two conjectures are false. The user can for instance apply Isabelle's counterexample checker to easily identify these.

\subsection{\tool's Neural Engine}
\label{sec:template-prediction}
\tool's neural engine uses a fine-tuned LLM to suggest templates given a theory context, which are then passed to \tool's symbolic engine (\S \ref{sec:symbolic-engine}).
For fine-tuning, we create labeled training data (input-output pairs) following \S \ref{sec:symbolic-engine}: we extract the template of each lemma appearing in a corpus of human-written theory files, and use a string representation of the template as the desired output. For each output template, its corresponding input is the list of symbols appearing in the body of the original lemma, as well as contextual information about the symbols. For a given symbol, both its type and any known associated definition convey information about how the symbol can be used. Although definitions provide more complete descriptions of symbols, types are more succinct and they often provide sufficient information to form well-typed templates and lemmas. Section \ref{sec:rq1} evaluates the effectiveness of types and definitions as inputs.
Revisiting the examples about octonions, the lemma \emph{octo\_distrib\_left} would result in the data-point of the shape:
\begin{align}
& \textit{Input: } [\textit{Symbols: } *, +] [\textit{Types of } *, +] [\textit{Defs of } *, +] \nonumber\\
& \textit{Output: } ?H_1\ x_1 \ (?H_2\ x_2\ x_3) = \label{eq:template_input} \\
& \hspace{2cm} ?H_2\ (?H_1\ x_1\ x_2)\  (?H_1\ x_1\ x_3) \nonumber
\end{align}
To obtain the list of symbols present in a given lemma, we first construct a theory file with low-level ML code to interact with Isabelle/Isar top-level and retrieve the lemma. To extract the symbols, we recursively iterate over applications in the term storing any encountered constants, using Isabelle-client~\citep{shminke2022pythonclientisabelleserver}. The list of symbols and their information is the minimal context necessary to recover a given lemma. However, additional context, such as relevant existing lemmas, may further guide the conjecturing process and is an interesting direction for future work.

\subsection{Direct Neural Conjecturing}
\label{sec:neural-baselines}
While \tool employs a neural engine to generate templates and a symbolic engine to instantiate those templates into lemmas, there is another approach one can take, which is to use a neural engine to directly generate lemmas. To create such a baseline, we adapt \tool's neural engine, by instead fine-tuning it on tuples of the form (symbols, context, lemma) instead of a template.
Revisiting the running example again, the lemma \emph{octo\_distrib\_left} now results in a data-point of the shape:
\begin{align}
& \textit{Input: } [\textit{Symbols: } *, +] [\textit{Types of } *, +] [\textit{Defs of } *, +] \nonumber \\
& \textit{Output: a * (b + c) = a * b + a * c} \label{eq:neural_input}
\end{align}

Two possible string representations of the lemma in the output are: (1) the \emph{lemma command} as appearing in the source code of the theory file (what a human would write), and (2) a string representation of the internal \emph{lemma object} within the proving engine. We noticed early in our experimentation that an LLM is almost never able to predict the lemma command as it is less structured than the lemma object, and so we use the lemma object representation. While there is some novelty in predicting lemma objects directly as opposed to lemma commands, the main purpose of doing direct neural lemma conjecturing is to create a suitable baseline for our experiments.

\subsection{Dataset}
\label{sec:dataset}
 We train and evaluate \tool on mathematical libraries from the Isabelle proof assistant. First, the Isabelle/HOL library, which contains formalizations based on higher order logic for a range of mathematics (e.g. number theory, analysis, algebra, set theory). Second, we also use the Isabelle Archive of Formal Proofs (AFP), which is a large collection of formalizations from the research community including mathematics, computer science and logic. At the time of writing, the AFP includes about 285,200 lemmas in close to 900 theories.
We extract templates for the lemmas in the above libraries as described in \S \ref{sec:symbolic-engine}.
In total, this results in a dataset of 62,816 data-points from the HOL library and 210,898 data-points from the AFP\footnote{Some theories could not be processed by the batch method we used, due to technical issues with theory imports. Hence, our dataset does not cover everything from the AFP.}.

%% file: main_contents/4_eval.tex
\section{Evaluation of \tool}\label{sec:experiments}
We set out to answer the following research questions:
\vspace{-0.5ex}
\begin{enumerate}[{RQ}1]
    \bfseries \item \mdseries How does our neuro-symbolic approach, \tool,
    compare with neural approaches and existing symbolic approaches? \vspace{-0.5ex}
    \bfseries \item \mdseries How does different contextual information (such as
    types and definitions) impact performance? \vspace{-0.5ex}
    \bfseries \item \mdseries What characterizes the conjectures \tool is able to generate (how many can be proved/have counter examples)? 
\vspace{-0.5ex}
\end{enumerate}

\subsection{Experimental Setup}

\paragraph{Models and Inference}
We evaluate three instances of \tool using DeepSeek-coder-1.3B, DeepSeek-coder-6.7B and Llama-3.2-1B as neural backends. These models are selected as representatives of open-weight base models performant on code-related tasks.   
One benefit of using smaller models in \tool is that it supports a more realistic setup for actual users wanting to use a conjecturing tool and run it locally on their machine. See Appendix~\ref{app:computing} for details about computing resources and replication parameters. 
At inference time, we experiment with both greedy decoding and beam search (size 4), plus a combination of the two.

\vspace{-1ex}
\paragraph{Baselines}

There are no existing neural-based tools or models for Isabelle to compare against so we train our own neural baselines (Section~\ref{sec:neural-baselines}), to directly suggest lemma. These are denoted by ``neural'' in tables in the evaluation. QuickSpec is the state-of-the-art symbolic tool for lemma conjecturing~\citep{quickspec}. QuickSpec is limited to generating equational lemmas as its conjecturing algorithm is based around enumerative synthesis of terms and the construction of equivalence classes via automated testing.

\vspace{-1ex}
\paragraph{Benchmarks}

We derive our datasets from the Isabelle/HOL library and the Archive of Formal Proofs (recall \S \ref{sec:dataset}) and create multiple train/validation/test sets from these libraries. 
We create two training sets: \textit{HOL-train} (57,576 datapoints containing only data from files in the HOL library) and \textit{HOL+AFP-train} (adding AFP projects published before 2024, resulting in 249,098 datapoints).
We also create two disjoint test sets: \textit{HOL-test} (4,740 datapoints) and \textit{AFP-test}
(18,975 datapoints from projects published in 2024). We note that while HOL-train and HOL-test share style and are about related topics, the data from the AFP tend to differ greatly in topic, content, and style, why we expect AFP-test to be more difficult.
We fine-tune our LLMs on the respective training sets and evaluate on both test sets to evaluate: in-context performance (HOL-train + HOL-test); the effect of adding more diverse training data (HOL+AFP-train + HOL-test); and finally the out-of-distribution performance (HOL-train/HOL+AFP-train + AFP-test). 
Evaluation on AFP-test also mitigates the risk of test data leakage for the fully neural baseline models\footnote{Recall that \tool's neural module produce templates rather than lemmas, and as the templates were created by us they are not present in prior training data.}, as their publicly reported pretraining cutoff dates are in 2023, before the AFP-test projects were published. 

For comparison with the symbolic tool QuickSpec, we separately test on the Octonions project~\citep{Octonions-AFP} from the AFP, which contains 350 lemmas, and leave it out of all training sets. We choose the Octonion formalization as it consists of equational lemmas, which is the domain QuickSpec supports. 


\vspace{-1ex}
\paragraph{Evaluation Metric}
A successful conjecturing system ought to produce lemmas that are \emph{interesting} and \emph{useful} in some sense. Note that what is interesting and useful is very much context dependent on for example existing definitions, lemmas and theorems, available tactics and potential different downstream tasks. Hence, there is non-trivial to define a metric that will be relevant across \emph{any} formalization in Isabelle. As a proxy, we therefore choose to evaluate against gold-standard lemmas written by humans in Isabelle, so they ought to align with what the human expert author thought was interesting and useful, and thus worth including in the formalization in the first place. 

In each evaluation task, there is one \emph{gold-standard lemma}. The method being assessed generates a set of suggested lemmas, and is considered successful if the gold-standard lemma is among these. 
Our matching criterion is based on Isabelle's internal representation of terms. For lemma-object predictions, we parse both the gold-standard lemma and each suggested lemma in Isabelle, in the source theory context, and compare their internal term representations.
This accepts harmless presentational differences such as alpha-renamed variables and eta-contracted forms. For template predictions, we similarly parse the predicted and gold templates and compare their Isabelle term representations. The metric is therefore stricter than proving equivalence: predictions that are merely logically equivalent after rewriting or other reasoning can still be counted as misses.

\vspace{-1ex}
\paragraph{Inference budget}
Note that the purely neural model and \tool are given the same neural GPU-inference budget per task. The neural models attempt to generate lemmas directly while \tool generates templates. \tool is then allowed an additional (symbolic) CPU-budget of max 60s per task to search for instantiations of the templates. \tool and the neural conjecturing model are also given the same input prompt, containing information about the symbols occurring in the relevant gold-standard target lemma (see Eq.~\ref{eq:template_input} in \ref{sec:template-prediction} and Eq.~\ref{eq:neural_input} in \ref{sec:neural-baselines}). These symbols are also used as the function name inputs to \tool's symbolic engine.

\subsection{RQ1--2: Discovery of Gold Standard Lemmas}
\label{sec:rq1}
\begin{table*}[t]
\caption{ Percentage of gold-standard lemmas discovered for \tool and a neural  version. We include results for the symbolic QuickSpec tool only on Octonions which is in its scope. Results are for 5 predictions (greedy decoding \& beam search with beam size 4). Note that due to limited computational resources and poor results using DeepSeek 1.3B, we did not further explore the prompting technique with only definitions for the other models.}
    \centering
    \begin{tabular}{lllllll}
& \multicolumn{3}{c}{HOL-train}  &  \multicolumn{3}{c}{HOL+AFP-train} \\ \hline
                 & HOL-test  & AFP-Test & Octonions & HOL-test  & AFP-Test & Octonions\\
                 [3pt] \hline \hline
DeepSeek-coder-1.3B \\ \hline
\tool (types + defs)     & \textbf{39.5\%} & \textbf{20.3\%} & 57.4\% & \textbf{41.8\%} & \textbf{28.2\%} & \textbf{66.9\%} \\
\tool (types)            & 37.6\% & 19.7\% & \textbf{58.3\%} & 35.7\% & 27.4\% & 56.0\% \\
\tool (defs)             & 33.2\% & 7.3\% & 54.6\% & 35.6\% & 13.0\% & 61.4\% \\
Neural (types + defs)    & 30.5\% & 11.9\% & 34.0\% & 30.9\% & 18.6\% & 46.0\% \\
Neural (types)           & 33.4\% & 12.9\% & 42.9\% & 30.1\% & 15.0\% & 48.0\% \\
Neural (defs)            & 30.5\% & 6.2\% & 41.4\% & 26.1\% & 7.8\% & 45.1\% \\
\hline
\tool Combined           & 48.0\% & 26.4\% & 67.7\% & 49.8\% & 35.9\% & 75.7\% \\
Neural Combined          & 42.6\% & 17.8\% & 53.1\% & 39.4\% & 23.7\% & 59.1\% \\
Combined                 & 53.3\% & 28.5\% & 70.9\% & 53.1\% & 38.7\% & 79.4\% \\[3pt] \hline \hline
DeepSeek-coder-6.7B \\ \hline
\tool (types + defs)     & \textbf{46.5\%} & \textbf{26.3\%} & \textbf{68.0\%} & \textbf{50.2\%} & 27.9\% & 76.6\% \\
\tool (types)            & 43.3\% & 22.5\% & 67.1\% & 44.5\% & \textbf{28.6\%} & \textbf{78.9\%} \\
Neural (types + defs)    & 42.3\% & 19.7\% & 53.1\% & 29.7\% & 25.1\% & 62.3\% \\
Neural (types)           & 38.6\% & 19.0\% & 51.4\% & 38.5\% & 25.2\% & 53.1\% \\
\hline
\tool Combined           & 52.2\% & 29.4\% & 76.0\% & 55.1\% & 34.8\% & 83.1\% \\
Neural Combined          & 46.8\% & 23.5\% & 59.4\% & 44.1\% & 31.6\% & 65.7\% \\
Combined                 & 57.2\% & 32.1\% & 79.7\% & 58.6\% & 42.7\% & 84.9\% \\[3pt] \hline \hline
Llama-3.2-1B \\ \hline
\tool (types + defs)     & 31.7\% & \textbf{17.7\%} & 57.4\% & 28.6\% & 19.3\% & 60.6\% \\
\tool (types)            & \textbf{34.1\%} & 15.0\% & \textbf{59.7\%} & \textbf{35.0\%} & \textbf{27.2\%} & \textbf{65.1\%} \\
Neural (types + defs)    & 23.1\% & 6.2\% & 40.6\% & 19.2\% & 12.9\% & 32.3\% \\
Neural (types)           & 22.8\% & 8.8\% & 42.3\% & 17.7\% & 12.2\% & 32.6\% \\
\hline
\tool Combined           & 39.3\% & 21.1\% & 62.6\% & 38.5\% & 30.4\% & 67.4\% \\
Neural Combined          & 28.9\% & 11.1\% & 48.6\% & 24.7\% & 17.7\% & 45.1\% \\
Combined                 & 42.9\% & 22.6\% & 64.0\% & 41.2\% & 32.9\% & 68.6\% \\[3pt]
\hline \hline
QuickSpec             &---               &---      & 22.8\% & --- & --- & 22.8\%  \\ 
 \bottomrule
\end{tabular}
\vspace{-1ex}
    
    
    \label{fig:rq1-main}
\end{table*}
To evaluate \tool and compare it against neural-only baselines and symbolic methods, we train \tool's neural model and the baseline models on either HOL-train or HOL+AFP-train, obtain multiple variants, and evaluate them across different test sets (Table~\ref{fig:rq1-main}). For each model, we compare the results obtained when various contextual information is included in the input: definitions, type information, or both (as described in \ref{sec:template-prediction}). Note that adding only the definitions, but not types, led to worse performance in initial experiments with DeepSeek 1.3B, so was not attempted for the other LLMs. For each model, we get 5 predictions: 1 using greedy decoding, and 4 using beam search (with beam size of 4). We break down the results for each decoding method in Appendix~\ref{app:greedy-results}.

We see that \tool outperforms the respective neural baselines on all test sets, with the best model (using DeepSeek 6.7B and HOL+AFP training) reaching 50.2\% on HOL-test and 28.6\% on AFP-test. For the smaller 1B models, the DeepSeek backend appears to consistently outperform Llama-3.2. 
We also see that neural methods are somewhat complementary to \tool: taken as an ensemble they discover even more lemmas, 58.6\% on HOL-test and 42.7\% on AFP-test (using DeepSeek 6.7B and HOL+AFP training). 
The inclusion of type information is greatly beneficial to both \tool and the neural baseline method.
We see that in some cases the success rate is higher when only type information is included and definitions are excluded, while in others including both is beneficial, although the differences are relatively small. 
We see that on AFP-test, the performance drops for all variants compared to their results on HOL-test. This is unsurprising, as the lemmas in AFP projects are more diverse than those in HOL. Also, the input to the LLMs may not always include enough contextual information for retrieving definitions in AFP theories requiring long-distance dependencies, though we always account for HOL as a dependency. 

For Octonions, \tool models (best reaching 78.9\%) outperform their respective neural baselines (62.3\%) while both greatly outperform QuickSpec (reaching just 22.8\%). 
Similar to other datasets, an ensemble of \tool and neural models provides a significant improvement, with overall lemma success rate up to 84.9\%. 
Not shown in Table~\ref{fig:rq1-main} is that QuickSpec generates several thousand lemmas, giving it an extremely poor precision of less than 1\%. This is because it has limited heuristics for judging which lemmas are interesting, and only skips lemmas that are logical consequences of existing lemmas. It also skips several useful and simple lemmas, such as $\mathit{inner}\, e_1\, x = \mathit{Im}_1\, x$, as they are considered trivial consequences of other lemmas. This is further discussed in Appendix \ref{app:quickspec}.

\begin{table*}[!t]
\caption{Gold-standard lemma discovery rates on the commercial LLM benchmark. Fine-tuned models use five predictions per task: one greedy prediction and four beam-search predictions, and the types+defs prompting strategy. Commercial LLMs also use five predictions per task: one greedy prediction and four temperature-1 samples.}
\centering
\setlength{\tabcolsep}{3.5pt}
\begin{tabular}{@{}lcc@{\hspace{1.4em}}lcc@{}}
\toprule
\multicolumn{3}{c}{\textbf{Template}} &
\multicolumn{3}{c}{\textbf{Lemma-object}} \\
\cmidrule(r){1-3}\cmidrule(l){4-6}
Model & Octonions & AFP-2026 &
Model & Octonions & AFP-2026 \\
\midrule
\tool DeepSeek-coder-1.3B & 66.9\% & 30.9\% &
Neural DeepSeek-coder-1.3B & 44.0\% & 25.2\% \\
\tool DeepSeek-coder-6.7B & \textbf{76.6\%} & \textbf{41.7\%} &
Neural DeepSeek-coder-6.7B & \textbf{61.7}\% & \textbf{38.1\%} \\
\tool Llama-3.2-1B & 46.9\% & 28.1\% &
Neural Llama-3.2-1B & 54.6\% & 23.0\% \\
Claude Opus 4.5 & 47.1\% & 19.4\% &
Claude Opus 4.5 & 46.3\% & 34.5\% \\
GPT-5.2 & 50.0\% & 15.8\% &
GPT-5.2 & 45.4\% & 25.2\% \\
\bottomrule
\end{tabular}
\vspace{-1ex}
\label{tab:commercial-llm-experiment}
\end{table*}




\subsection{RQ3: Analysis of Generated Conjectures}
\label{sec:qualitative}
In \S \ref{sec:rq1}, we counted the number of gold standard lemmas discovered. Here, we break down these results further.

\paragraph{Applicability to Realistic Formalizations}
Appendix \ref{app:breakdown-theories}
shows a breakdown per projects/theories in HOL-test and 
AFP-test, with the number of test lemmas from each and the success rate of the best-performing \tool and neural ensembles with DeepSeek 6.7B. We note that \tool performs better on almost every test theory in the test sets. 
We also list the topic of the different AFP formalizations in our test set, showing that they range over a variety of different topics from Computer Science, Mathematics and Logic. The results confirm that \tool's approach to conjecturing by analogy successfully produces interesting lemmas across a wide range of real world formalizations. 

\paragraph{Generated Conjectures}
We take a closer look at the full set of conjectures generated by the best-performing instance of \tool (DeepSeek 6.7B and HOL+AFP training), under greedy decoding. 
We separate the generated conjectures into three main categories (true, false and unresolved) using the Sledgehammer tool for automated proofs \cite{sledgehammer}, and counter-example finding tools  Nitpick \cite{nitpick} and Quickcheck \cite{quickcheck}. \emph{True conjectures} consist of 1) the gold-standard lemmas, and 2) other conjectures proved by Sledgehammer. \emph{False conjectures} are those for which we can find a counter-examples. \emph{Unresolved conjectures} are those that can neither be proved by Sledgehammer, nor have a counter-example. 

For HOL-test (4,740 gold-standard lemmas), \tool generates 7,628 conjectures.
Table \ref{tab:novelty} shows a breakdown of these conjectures: 1,356 are  true statements which don't match the gold-standard lemmas, and 2,749 are unresolved conjectures which require further non-trivial reasoning. Appendix \ref{app:qualitative} contains some examples of these, and Appendix \ref{app:statistics} additional statistics.

\begin{table}[h]

\caption{Generated conjectures for HOL-test.
}
\centering
\begin{tabular}{@{}llll@{}}
\toprule
Category    & Size \\ \midrule
Gold-standard & 1721 (22.6\%)\\
Other True Statements & 1356 (17.8\%) \\
False Conjectures & 1802 (23.6\%) \\
Unresolved Conjectures & 2749 (36.0\%) \\
\textbf{Total} & 7628 \\
 \bottomrule
\end{tabular}

    \label{tab:novelty}

\vspace{-2mm}
\vskip -0.1in
\end{table}

\section{Comparison with Commercial LLMs}
\label{sec:commercial}
A common question arising is whether a commercial closed-source LLM would have been able to perform the conjecturing task as well or better without any additional fine tuning: 

\textbf{RQ4: How does large commercial instruction fine-tuned models perform both as backends for \tool, and on the direct conjecturing task?} 

\noindent As a complement to our results using auto-regressive fine-tuned models above, we therefore also evaluate Claude Opus 4.5 and GPT-5.2 with few-shot learning.  
For economic reasons, this evaluation is conducted on a limited test set: our running example of Octonions, as well as two formalizations added to Isabelle's AFP in 2026 involving the common list datatype: \emph{Swap Distance} \cite{Swap_Distance-AFP}, and \emph{Linear Orders as Rankings} \cite{Rankings-AFP}. The training cut-offs for both Claude Opus 4.5 and GPT-5.2 are in 2025, so using two very recent formalizations ought to minimize\footnote{Development versions of these formalizations might of course have been on the internet prior to 2026, so we cannot fully guarantee there was no leakage.} the probability of training data leakage.

While the setup is different between instruction fine-tuned commercial models and the auto regressive fine-tuned models in the previous section, we attempt to compare them under as similar conditions as possible. We used five few-shot examples per problem, both for template generation and for direct conjecture generation. Relevant examples are selected from our AFP-train dataset to reflect a reasonably fair comparison. Details of this selection as well as the prompts to the LLMs are in Appendix \ref{app:commercial}. The success rates for the commercial models are also calculated based on the top-5 predictions. Note however that the commercial models consume considerably more compute and tokens: for comparison the inputs for template prediction for the fine-tuned models are about 325 token per query, while for the commercial models it is about 1700 tokens.
The results are summarised in Table \ref{tab:commercial-llm-experiment}. 

%% file: main_contents/5_conclusion.tex
\section{Conclusion}
\tool is a novel neuro-symbolic bottom-up conjecturing tool for the Isabelle proof assistant. \tool works by 
(neurally) learning analogical families of conjectures, represented as templates which are symbolically instantiated in new theories.
This approach is efficient at discovering lemmas in formalizations across a wide range of topics from Isabelle's HOL and AFP libraries. \tool outperforms fine-tuned neural baselines, symbolic tools and, on a test set including recent formalizations, also commercial models Claude Opus 4.5 and GPT-5.2. We demonstrate that by fine-tuning smaller models on the simpler task of template generation, good results are achievable with a modest amount of compute.

The best-performing \tool backend (deepseek-6.7b), discovers 50\% of gold standard lemmas for HOL-test, 29\% for AFP-test and 41\% on AFP-2026. 
Ensembling different prompting strategies further improves results, showing that the different approaches discover slightly different conjectures.

This work is a first step toward demonstrating the usability of neuro-symbolic analogical conjecturing for proof assistants on a realistic selection of theories from real formalizations in computer science, mathematics, and logic, going beyond benchmark theories from high-school and undergraduate mathematics competitions. 



%% file: Appendices/template_language.tex
\section{Template language and symbols}
\label{app:templates}
Templates are implemented as instances of Isabelle/HOL's term datatype\footnote{See section 2.2 of the Isabelle/Isar Implementation Manual \url{https://isabelle.in.tum.de/dist/Isabelle2025/doc/implementation.pdf}}. 
While theory-specific symbols are replaced by holes, general logical symbols remain in the template to not make it overly abstract and obscuring the analogies we wish to uncover. The symbols that are part of the template language and not abstracted away are:
\begin{itemize}
    \item All constants whose names begin with ‘HOL.’  These are the functions defined in Isabelle/src/HOL/HOL.thy including equality, True, False, Not, All and Ex (quantifiers), conjunction and disjunction.
    \item All constants whose names begin with ‘Pure.’  Basic logical constructs including implication (Pure.imp), Pure.all, Pure.eq, defined in Isabelle/src/Pure/logic.ML
    \item Bounded quantifiers for sets (these are rendered the same as the regular quantifiers defined in HOL).
    \item Set membership.
    \item Pairs/Cartesian products as defined in src/HOL/Product\_Type.
    \item Inequality symbols: less, greater, less or equal, greater or equal
  Defined in "Orderings.ord\_class.less\_eq" and "Orderings.ord\_class.less"
  (greater and greater\_eq are defined in terms of less and less\_eq and are translated in the term structure so we don’t expect them to appear in templates).
\end{itemize}
Implementation details of the exact symbols and how to extract them are available in our anonymous code repository.

%% file: Appendices/resources.tex
\section{Computing Resources and Details for Replication}
\label{app:computing}
We run the majority of our training and evaluation on 8 NVIDIA RTX 2080 Ti (11GB VRAM). The Deepseek 6.7B models are trained and evaluated on 4 NVIDIA Tesla A40s, each with 48GB of VRAM. We use 8-bit quantization when loading models for both fine-tuning and inference. We train models with DistributedDataParallel (DDP) for 12 epochs and $8e-4$ learning rate. We use a maximum sequence length of 1024 with 200 tokens reserved for the output. We use an effective batch size 16.

%% file: Appendices/greedy_results.tex
\section{Results found using greedy decoding vs beam search}
\label{app:greedy-results}

\begin{table*}[!t]
    \centering
    \caption{Breakdown of Table~\ref{fig:rq1-main} by decoding strategy. Top: greedy decoding. Bottom: beam search with beam size 4.} 
    \small
    \newcommand{\appdoublecline}{\cline{2-8}\noalign{\vskip\doublerulesep}\cline{2-8}}
    \resizebox{\textwidth}{!}{
    \begin{tabular}{@{}clllllll@{}}
\hline
\multirow{29}{*}{\rotatebox[origin=c]{90}{\textbf{Greedy decoding}}}
&                        & \multicolumn{3}{c}{HOL-train} & \multicolumn{3}{c}{HOL+AFP-train} \\ \cline{2-8}
& Method                 & HOL-test & AFP-Test & Octonions & HOL-test & AFP-Test & Octonions \\ \appdoublecline
& DeepSeek-coder-1.3B    \\ \cline{2-8}
& \tool (types + defs)   & \textbf{28.7\%} & 14.7\% & 47.1\% & \textbf{29.1\%} & \textbf{20.0\%} & \textbf{54.0\%} \\
& \tool (types)          & 27.3\% & \textbf{15.1\%} & \textbf{50.0\%} & 24.4\% & 19.9\% & 42.9\% \\
& \tool (defs)           & 23.8\% & 4.8\% & 45.1\% & 24.3\% & 8.1\% & 53.1\% \\
& Neural (types + defs)  & 23.5\% & 8.8\% & 30.0\% & 26.1\% & 15.4\% & 39.7\% \\
& Neural (types)         & 20.5\% & 9.8\% & 29.4\% & 24.1\% & 11.6\% & 42.9\% \\
& Neural (defs)          & 22.8\% & 3.9\% & 31.4\% & 20.8\% & 5.2\% & 36.9\% \\ \cline{2-8}
& \tool Combined         & 37.3\% & 21.0\% & 58.3\% & 37.5\% & 27.3\% & 61.7\% \\
& Neural Combined        & 32.6\% & 13.7\% & 44.9\% & 33.5\% & 19.7\% & 51.1\% \\
& Combined               & 42.9\% & 23.5\% & 62.0\% & 43.3\% & 31.0\% & 66.6\% \\[3pt] \appdoublecline
& DeepSeek-coder-6.7B    \\ \cline{2-8}
& \tool (types + defs)   & \textbf{34.9\%} & \textbf{19.8\%} & \textbf{60.9\%} & \textbf{36.6\%} & 4.7\% & \textbf{62.6\%} \\
& \tool (types)          & 31.6\% & 18.2\% & 55.7\% & 31.2\% & 7.3\% & 57.7\% \\
& Neural (types + defs)  & 30.9\% & 14.2\% & 46.9\% & 16.2\% & 3.5\% & 33.7\% \\
& Neural (types)         & 29.2\% & 14.6\% & 42.9\% & 8.8\% & \textbf{18.9\%} & 48.0\% \\ \cline{2-8}
& \tool Combined         & 40.5\% & 23.3\% & 66.3\% & 42.4\% & 10.2\% & 70.0\% \\
& Neural Combined        & 36.0\% & 18.7\% & 50.9\% & 18.0\% & 20.1\% & 53.7\% \\
& Combined               & 45.8\% & 26.7\% & 70.3\% & 44.4\% & 23.2\% & 74.9\% \\[3pt] \appdoublecline
& Llama-3.2-1B           \\ \cline{2-8}
& \tool (types + defs)   & 16.3\% & 9.8\% & \textbf{44.0\%} & 17.1\% & 12.3\% & 46.0\% \\
& \tool (types)          & \textbf{17.9\%} & \textbf{9.9\%} & \textbf{44.0\%} & \textbf{19.4\%} & \textbf{15.2\%} & \textbf{49.7\%} \\
& Neural (types + defs)  & 12.0\% & 3.6\% & 23.4\% & 9.9\% & 5.6\% & 12.9\% \\
& Neural (types)         & 11.6\% & 4.4\% & 23.1\% & 11.1\% & 7.1\% & 20.6\% \\ \cline{2-8}
& \tool Combined         & 23.3\% & 13.8\% & 52.0\% & 24.5\% & 19.1\% & 54.3\% \\
& Neural Combined        & 17.0\% & 6.2\% & 33.4\% & 15.7\% & 9.9\% & 26.9\% \\
& Combined               & 27.6\% & 15.3\% & 54.3\% & 27.9\% & 21.2\% & 57.1\% \\[3pt] \cline{2-8}
\multirow{29}{*}{\rotatebox[origin=c]{90}{\textbf{Beam search}}}
&                        & \multicolumn{3}{c}{HOL-train} & \multicolumn{3}{c}{HOL+AFP-train} \\ \cline{2-8}
& Method                 & HOL-test & AFP-Test & Octonions & HOL-test & AFP-Test & Octonions \\ \appdoublecline
& DeepSeek-coder-1.3B    \\ \cline{2-8}
& \tool (types + defs)   & \textbf{37.6\%} & \textbf{19.1\%} & \textbf{52.6\%} & \textbf{40.0\%} & \textbf{26.6\%} & \textbf{63.7\%} \\
& \tool (types)          & 35.8\% & 18.1\% & 49.7\% & 33.5\% & 25.9\% & 52.9\% \\
& \tool (defs)           & 31.8\% & 6.8\% & 49.1\% & 33.5\% & 12.3\% & 58.3\% \\
& Neural (types + defs)  & 29.3\% & 11.5\% & 31.7\% & 27.9\% & 16.7\% & 45.4\% \\
& Neural (types)         & 33.0\% & 12.5\% & 42.3\% & 28.0\% & 14.1\% & 44.9\% \\
& Neural (defs)          & 30.2\% & 6.2\% & 41.4\% & 24.0\% & 7.3\% & 41.7\% \\ \cline{2-8}
& \tool Combined         & 46.8\% & 25.5\% & 66.0\% & 48.2\% & 34.4\% & 74.6\% \\
& Neural Combined        & 42.1\% & 17.3\% & 52.9\% & 37.3\% & 22.2\% & 57.4\% \\
& Combined               & 52.6\% & 27.9\% & 69.4\% & 51.7\% & 37.2\% & 79.1\% \\[3pt] \appdoublecline
& DeepSeek-coder-6.7B    \\ \cline{2-8}
& \tool (types + defs)   & \textbf{46.4\%} & \textbf{26.2\%} & \textbf{68.0\%} & \textbf{50.1\%} & 27.8\% & 76.6\% \\
& \tool (types)          & 43.2\% & 16.9\% & 66.9\% & 36.2\% & \textbf{28.4\%} & \textbf{78.3\%} \\
& Neural (types + defs)  & 42.3\% & 19.7\% & 53.1\% & 22.3\% & 25.0\% & 62.3\% \\
& Neural (types)         & 38.4\% & 18.9\% & 51.1\% & 38.4\% & 24.9\% & 53.1\% \\ \cline{2-8}
& \tool Combined         & 52.1\% & 28.9\% & 76.0\% & 54.2\% & 34.7\% & 83.1\% \\
& Neural Combined        & 46.8\% & 23.5\% & 59.4\% & 43.2\% & 31.4\% & 65.7\% \\
& Combined               & 57.1\% & 31.9\% & 79.7\% & 57.9\% & 42.5\% & 84.9\% \\[3pt] \appdoublecline
& Llama-3.2-1B           \\ \cline{2-8}
& \tool (types + defs)   & 31.3\% & \textbf{17.4\%} & 57.4\% & 27.3\% & 17.6\% & 58.9\% \\
& \tool (types)          & \textbf{33.7\%} & 13.6\% & \textbf{59.7\%} & \textbf{34.2\%} & \textbf{26.8\%} & \textbf{64.9\%} \\
& Neural (types + defs)  & 22.6\% & 5.4\% & 40.0\% & 18.8\% & 12.5\% & 32.0\% \\
& Neural (types)         & 22.3\% & 8.5\% & 42.0\% & 15.7\% & 11.1\% & 27.4\% \\ \cline{2-8}
& \tool Combined         & 38.8\% & 20.5\% & 62.6\% & 37.6\% & 29.5\% & 66.6\% \\
& Neural Combined        & 28.3\% & 10.5\% & 48.0\% & 23.4\% & 16.8\% & 43.4\% \\
& Combined               & 42.1\% & 21.9\% & 64.0\% & 40.1\% & 32.0\% & 68.3\% \\[3pt] \cline{2-8}
& QuickSpec              & --- & --- & 22.8\% & --- & --- & 22.8\% \\
\bottomrule
\end{tabular}}
    \label{fig:decoding-results}
\end{table*}

We break down the results from Table~\ref{fig:rq1-main} in Table~\ref{fig:decoding-results} to show the performance of different decoding strategies. The results in Table~\ref{fig:decoding-results} are for greedy decoding and beam search. Greedy decoding selects the token with the highest probability at each generation step. Beam search obtains multiple high-probability predictions for a given model. We see that beam search helps to improve results for both \tool and the neural baseline ensembles. However, \tool is still performant in a greedy setting, showing that the approach can achieve good results with a smaller inference budget.

%% file: Appendices/HOL_results.tex
\subsection{HOL-test}
\label{app:hol-test}

\begin{table*}[!t]
\caption{Results for the different sessions of the HOL library.}
\label{hol-results}
\vskip 0.15in
    \centering
\small
\begin{tabular}{l|rrrr}
\toprule
\textbf{HOL session} & \textbf{Lemmas} & \textbf{\tool} & \textbf{Neural} & \textbf{Combined}\\
\midrule
Algebra            & 49 & \textbf{73.47\%}  & 67.35\%  & 81.63\%  \\
Analysis & 1009 & \textbf{41.72\%}  & 28.84\%  & 44.80\%  \\
Auth               & 350 & \textbf{61.71\%}  & 60.00\%  & 64.29\%  \\
Bali            & 346 & \textbf{40.46\%}   & 21.10\%  & 41.62\%  \\
Cardinals        & 93 & \textbf{46.24\%}  & 34.41\%  & 46.24\%  \\
Computational\_Algebra & 50 & \textbf{66.00\%}  & 62.00\%  & 66.00\%  \\ 
Data\_Structures & 91 & \textbf{74.73\%}  & 70.33\%  & 76.92\%  \\
Datatype\_Examples   & 145 & \textbf{53.10\%} & 23.45\%  & 55.17\%  \\
Decision\_Procs & 240 & \textbf{58.75\%}  & 47.08\%  & 60.42\%  \\
Eisbach          & 23 & 0.00\% & 0.00\% & 0.00\% \\
ex               & 75 & \textbf{41.33\%}  & 37.33\%  & 42.67\%  \\
Hoare            & 15 & \textbf{80.00\%} & 66.67\% & 80.00\%  \\
HOLCF            & 93 & \textbf{61.29\%}  & 47.31\%  & 63.44\%  \\ 
IMP             & 284 & \textbf{89.08\%}  & 88.73\%  & 95.07\%  \\
Imperative\_HOL  & 61 & \textbf{40.98\%}  & 27.87\%  & 45.90\%  \\
IMPP             & 26 & \textbf{50.00\%}  & 26.92\%  & 53.85\%  \\
IOA              & 20 & \textbf{45.00\%}  & 35.00\%  & 50.00\%  \\
Lattice              & 55 & \textbf{61.82\%}   & 54.55\%  & 69.09\%  \\
Library            & 1054 & \textbf{54.74\%} & 39.47\%  & 58.25\%  \\
Matrix\_LP          & 25 & \textbf{24.00\%}  & 20.00\%  & 24.00\%  \\
Metis\_Examples   & 1 & 0.00\%  & 0.00\%  & 0.00\%  \\
MicroJava       & 178 & \textbf{47.75\%}  & 34.27\%  & 51.12\%  \\
Nominal         & 302 & \textbf{81.79\%}  & 61.26\%  & 84.44\%  \\
Nonstandard\_Analysis & 36 & \textbf{75.00\%}  & 72.22\%  & 80.56\%  \\
Number\_Theory   & 47 & \textbf{61.70\%}  & \textbf{61.70\%}  & 72.52\%  \\ 
Quotient\_Examples  & 18 & \textbf{55.56\%} & 38.89\%  & 55.56\%  \\
Proofs            & 6 & \textbf{33.33\%}  & 16.67\%  & 33.33\%  \\
TLA               & 7 & \textbf{57.14\%}  & 28.57\%  & 57.14\%  \\
UNITY            & 41 & 14.63\%  & \textbf{19.51\%}  & 24.39\%  \\ 
\bottomrule
\end{tabular}
\end{table*}

For the lemmas included in our test set HOL-test, Table~\ref{hol-results} shows the number of test lemmas from each session and the lemma success rates (in percentages) of \tool, neural-only lemma prediction, and their combination. The results shown are the ensembles of the different variants of \tool and the neural baselines, trained on HOL+AFP-train using DeepSeek-coder-6.7b. We created a file-wise split of the HOL library.

For the 29 different sessions, we see that the \tool ensemble outperforms the neural ensemble on 25 of them, while the neural ensemble performs better in one case, and in the remaining three both methods get the same rate of accurate predictions (0\% in two of the three). On average the success rate of the \tool ensemble is 10.34\% higher than that of the neural ensemble. Their combined success rate is on average 3.00\% higher than that of the \tool ensemble, showing that the approaches are somewhat complementary but mostly overlapping.

A clear outlier is the Eisbach session where both methods are completely unsuccessful, this is because the lemmas in that set are all simple example lemmas used to demonstrate the use of the Eisbach proof method language~\cite{matichuk2016eisbach}. These examples \footnote{\url{https://isabelle.in.tum.de/library/HOL/HOL-Eisbach/Examples.html}} only contain basic logic symbols which we discard from our input symbol set, so our models don't have any input to base their predictions on.

%% file: Appendices/AFP_results.tex
\subsection{AFP-test}
\label{app:afp-test}
For the formalizations included in our test set AFP-test, Table~\ref{afp-results} shows the formalization topic chosen by the project authors, the number of gold-standard lemmas from each project and the lemma success rates (in percentages) of \tool, neural-only lemma prediction, and their combination.  The results shown are the ensembles of the different variants of \tool and the neural baselines, trained on HOL+AFP-train using DeepSeek-coder-6.7b.

We can see that out of the 27 tested formalization projects, the \tool ensemble outperforms the neural ensemble on 19 of them, while the neural ensemble performs better for two, and in the remaining 6 cases the performance is equally good (0\% in five out of six cases). Note that most of the projects with a success rate of 0\% contain a small number of target lemmas (less than 10), with the outlier being the Actuarial Mathematics formalization. In this case, the syntax used in the formalization meant that none of the predicted lemmas were considered syntactically equivalent, however a more thorough evaluation would have shown some of them to be semantically equivalent to the target lemmas. On average the \tool ensemble has a 7.18\% better performance than the neural one. Their combined success rate is on average 3.52\% higher than the one for the \tool ensemble. 

\begin{table*}[!t]
\caption{Information about the different formalization projects in AFP-test.}
\label{afp-results}
\vskip 0.15in
    \centering
\small

\begin{tabular}{ll|crrr}
\toprule
\textbf{AFP entry} & \textbf{Topic(s)} & \textbf{Lemmas} & \textbf{\tool} & \textbf{Neural} & \textbf{Combined}\\
\midrule
ConcurrentHOL         & CS/Concurrency            & 481 & \textbf{39.71\%} & 28.27\% & 43.24\% \\
\hline
Sumcheck & CS/Security               & 62  & \textbf{67.74\%} & 62.90\% & 74.19\% \\
Protocol     & CS/Algorithms &     &       &       &       \\
\hline
Broadcast\_Psi & CS/Concurrency            & 145 & \textbf{82.07\%} & 69.66\% & 85.52\% \\
\hline
AutoCorres2           & CS/PL  & 12721 & \textbf{34.72\%} & 22.48\% & 37.86\% \\
                      & CS/Semantics \& reasoning &     &       &       &       \\
                      & Tools &     &       &       &       \\    
\hline
Substitutions & Logic/Rewriting & 55 & \textbf{50.91\%} & 21.82\% & 52.73\% \\
Lambda-Free &&&&\\
\hline
Doob\_Convergence & Math/Prob. theory & 2 & 0.00\% & 0.00\% & 0.00\% \\   
\hline
Orient\_Rewrite & Logic/Rewriting & 148 & \textbf{60.14\%} & 42.57\% & 62.16\% \\
Rule\_Undecidable &&&&\\
\hline
Uncertainty & Math/Physics & 3 & 0.00\% & 0.00\% & 0.00\% \\   
Principle &&&&&\\
\hline
Derandomization    & CS/Algorithms &15 & 40.00\% & 40.00\% & 40.00\% \\ 
with Conditional &&&&\\
Expectations &               &   &       &       &       \\
\hline
Verified QBF & CS/Algorithms & 171 & \textbf{53.80\%} & 43.86\% & 57.31\% \\
Solving                        & Logic/General logic &     &       &       &       \\
\hline
IMP & CS/PL & 50 & \textbf{40.00\%} & 38.00\% & 42.00\% \\
Noninterference                     & CS/Security &     &       &       &       \\ 
\hline
Actuarial & Math/Games \& econ. & 168 & 0.00\% & 0.00\% & 0.00\% \\                
 Mathematics  &&&&\\
 \hline
LL(1) Parser & CS/Algorithms & 205 & 38.05\% & \textbf{46.83\%} & 55.61\% \\       
Generator                     & CS/PL &     &       &       &       \\
\hline
Schönhage-Strassen & CS/ALgorithms & 128 & \textbf{41.41\%} & 26.56\% & 43.75\% \\
Multiplication                     & Math/Algebra &     &       &       &       \\
\hline
Isabelle\_DOF & CS/Semantics \& reasoning & 41 & \textbf{53.66\%} & 34.15\% & 53.66\% \\
\hline
Interval Analysis & Math/Analysis & 694 & \textbf{55.19\%} & 46.83\% & 58.50\% \\
\hline
MFOTL Checker & CS/Data mgmt systems & 2081 & 14.22\% & \textbf{24.89\%} & 30.71 \% \\    
          & CS/Algorithms &    &       &       &       \\                              
                    & Logic/General logic &    &       &       &       \\
\hline
Decomposition of & Math/Algebra & 1 & 0.00\% & 0.00\% & 0.00\%\\           
totally ordered hoops &&&&\\
\hline
Approximate Model & CS/Algorithms & 112 & \textbf{32.14\%} & 17.86\% & 34.82\% \\        
Counting &&&&&\\
\hline
Wieferich-Kempner & Math/Number theory & 8 & 0.00\% & 0.00\% & 0.00\% \\  
Theorem &&&&\\
\hline
PNT\_with & Math/Number theory & 155 & \textbf{35.48\%} & 25.81\% & 40.00\% \\
Remainder &&&&\\
\hline
Continued Fractions & Math/Analysis & 428 & \textbf{37.62\%} & 35.75\% & 43.69\% \\
\hline
CondNormReasHOL & Logic/Phil. aspects & 34 & \textbf{11.76\%} & 5.88\% & 11.76\% \\
                & Logic/General logic &    &       &       &       \\
\hline
Region Quadtrees & CS/Data structures & 182 & \textbf{41.76\%} & 34.07\% & 46.15\% \\
\hline
Karatsuba & CS/Algorithms & 431 & \textbf{50.58\%} & 32.25\% & 55.22\% \\
\hline
Pick's Theorem & Math/Geometry & 334 & \textbf{19.46\%} & 14.37\% & 22.46\% \\
\hline
Kummer Congruence & Math/Number theory & 120 & \textbf{34.17\%} & 25.83\% & 38.33\% \\
\bottomrule
\end{tabular}
\end{table*}

%% file: Appendices/lemma_statistics.tex
\section{Additional Test Set Statistics}
\label{app:statistics}

\begin{table*}[!t]
\centering
\caption{ Number of lemmas in each test set, and how many were discovered by \tool and the corresponding neural method (using DeepSeek-6.7b and greedy decoding), including how many were equational. The table also shows information about template and lemma lengths (number of characters) and the number of different symbols appearing.
}
\resizebox{\textwidth}{!}{
\begin{tabular}{@{}lllllllllllllll@{}}
\toprule
\multirow{2}{*}{} & \multirow{2}{*}{Lemmas} & \multirow{2}{*}{Eq} & \multicolumn{5}{c}{Template Length} & \multicolumn{5}{c}{Lemma Length}   & \multicolumn{2}{c}{\# Symbols} \\ \cmidrule(l){4-15} 
                  &                              &                             & Min  & 25\%  & Med& 75\% & Max  & Min & 25\% & Med & 75\% & Max   & Mean           & Max                   \\ [3pt] \hline \hline
HOL-test  & 4740  & 2242  & 9 & 49 & 82 & 127 & 9014 & 7 & 57 & 91 & 150 & 10454 & 4.2 & 100 \\ \hline
\tool & 1643 & 885 & 11 & 37 & 55 & 85 & 300 & 10 & 45 & 69 & 100 & 368 & 3.3 & 15 \\
Neural & 1407  & 750 & 11 & 37 & 55 & 85 & 433 & 11 & 45 & 68 & 98 & 390 & 3.3 & 15 \\ [3pt] \hline \hline
AFP-test & 18975 & 13202 & 7 & 39 & 63 & 130 & 10492 & 7 & 58 & 80 & 141 & 7530 & 4.1 & 31 \\ \hline
\tool & 3712 & 2989 & 7 & 25 & 31 & 45 & 220 & 12 & 39 & 55 & 73 & 286 & 2.8 & 14 \\
Neural & 2111 & 1526 & 7 & 23 & 33 & 45 & 184 & 10 & 36 & 52 & 71 & 253 & 2.8 & 12 \\ [3pt] \hline \hline
Octonions & 350 & 262 & 17 & 29 & 49 & 75 & 688 & 11 & 33 & 55 & 79 & 391 & 3.8 & 21 \\ \hline
\tool & 212 & 152 & 17 & 23 & 43 & 67 & 121 & 11 & 29 & 47 & 65 & 133 & 2.7 & 11 \\
Neural & 164 & 105 & 19 & 23 & 49 & 67 & 269 & 15 & 25 & 47 & 66 & 279 & 2.7 & 11.0 \\ \bottomrule
\end{tabular}}

\label{fig:statistics}
\end{table*}

Statistics on HOL-test, AFP-test, and Octonions, and the successes of \tool and neural baselines on each dataset are shown in Table~\ref{fig:statistics}. We count the number of lemmas in each test set (Lemmas), how many of them are equational (Eq), and look into the length (in characters) of lemmas and their templates as well as the number of different symbols that appear. \tool and neural models here are trained on HOL-train and use greedy decoding, using DeepSeek-coder-6.7b for both the neural prediction and the neural engine of \tool.

%% file: Appendices/qualitative_results.tex
\section{Examples of Generated Conjectures}\label{app:qualitative}




The analysis presented in section \ref{sec:qualitative} divides the set of conjectures generated by \tool into \emph{true}, \emph{false}, and \emph{undecided} conjectures. Here, we present some examples of conjectures in each category.

\paragraph{True Statements} If Sledgehammer succeeds in finding a proof of a given conjecture, it's considered a true statement. True statements included both gold-standard lemmas formalized by experts and other other true statements not previously formalized. The following two conjectures were discovered by \tool: the first lemma  is identical to the gold-standard lemma $totient\_imp\_prime$ and the second one doesn't match any formalized lemmas.

\begin{lemma}
    \(``totient\ x_1 = x_1 - 1 \Longrightarrow 0 < x_1 \Longrightarrow prime\ x_1"\)
\end{lemma}
\begin{lemma}
    \(``1 \le x_1 \Longrightarrow 1 \le totient\ x_1"\)
\end{lemma}

\paragraph{False Statements} Conjectures for which Nitpick or Quickcheck can find a counter-example are considered false statements. For the conjecture below, Quickcheck finds a counter-example when $x_1 = 2$.

\begin{lemma}
 \(``2 \le x_1 \Longrightarrow even\ (totient\ x_1)"\)
 \end{lemma}

\paragraph{Unresolved Conjectures} \tool generates a number of conjectures which can't be automatically categorized. This is expected due to the incompleteness of the automated reasoning tools we use. These conjectures require further complex reasoning to decide if that are true or false. One example of such an unresolved conjecture generated by \tool is the monotonicity of Euler's function. Nitpick and Quickcheck fail to find a counterexample such as $x_1 = 21, x_2=22$.

\begin{lemma}
\(``x_1 \le x_2 \Longrightarrow totient\ x_1 \le totient\ x_2"\)
\end{lemma}

%% file: Appendices/commercial.tex
\section{Comparison with Commercial LLMs Details}
\label{app:commercial}

\subsection{Evaluation Sets}
We evaluate commercial models on Octonions and on two fresh 2026 AFP entries, Rankings and Swap Distance. 
The Octonions benchmark is identical to the one used for evaluating RQ1-2, and it contains 350 datapoints.
The Rankings and Swap Distance benchmark contains 139 datapoints extracted from three source theories in an AFP snapshot from 2026-04-26, using Isabelle2025-2.
For each datapoint, the extracted record contains the gold-standard lemma object, its abstracted template, and the symbol, type, and definition information used to construct model inputs.

\subsection{Prompt Format}
Each prompt contains an instruction, five in-context examples, and a final query.
Each example provides the available symbols, relevant definitions, and the desired output.
The query has the same symbol and definition fields but leaves the output blank for the model to complete.

The full prompt has the following form:

\begin{lstlisting}[basicstyle=\small\ttfamily,breaklines=true]
<task instruction>

### Example <i>
###symbols
<symbol table for example>
###defs
<definitions for example>
###output
<gold output for example>###end

... repeated for five in-context examples ...

### Now predict
###symbols
<query symbol table>
###defs
<query definitions>
###output
\end{lstlisting}

For template prediction, the instruction is:

\begin{lstlisting}[basicstyle=\small\ttfamily,breaklines=true]
You are predicting Isabelle/HOL lemma templates for the Lemmanaid task.

You will be given a context block describing the symbols (with type signatures) and definitions available, then asked to produce a template -- an abstracted Isabelle proof goal where concrete operators are replaced by placeholders (?H1, ?H2, ...) and bound variables by x_1, x_2, .... 

The symbol table determines the placeholder inventory exactly. If there are N listed symbols, your output must use exactly the placeholders ?H1 through ?HN. Every one of ?H1, ?H2, ..., ?HN must appear at least once in the final template. No other ?H placeholders may appear. 

Below are worked examples from the same theory. Then a new context. Output ONLY the template followed by `###end`. No prose, no markdown fences.
\end{lstlisting}

For direct lemma-object prediction, the instruction is:

\begin{lstlisting}[basicstyle=\small\ttfamily,breaklines=true]
You are predicting Isabelle/HOL lemma statements for the Lemmanaid task.

You will be given a context block describing the symbols (with type signatures) and definitions available, then asked to produce the lemma statement -- a concrete Isabelle proof goal using the named operators from the symbol table and schematic variables (?x, ?y, ...). 

The symbol table determines the required operator inventory exactly. Every listed symbol must be used at least once in the final lemma statement, either by name or by its standard Isabelle notation when appropriate. Do not use operators that are not listed in the symbol table, except built-in logical syntax and schematic variables. 

Below are worked examples from the same theory. Then a new context. Output ONLY the lemma statement followed by `###end`. No prose, no 
markdown fences.
\end{lstlisting}

\subsection{In-context Examples}
The prompt generator supports several strategies for selecting in-context examples.
The results reported in Table~\ref{tab:commercial-llm-experiment} use nearest-example selection.
The examples are drawn from a fixed AFP-train pool to simulate a realistic conjecturing setting.
Nearestness is measured by overlap in the Isabelle symbols appearing in the task: for a query with symbol set $Q$, each candidate example with symbol set $S$ is scored by Jaccard overlap, $\frac{|Q \cap S|}{|Q \cup S|}$.

%% file: Appendices/qs_octonions.tex
\section{Partial QuickSpec output on Octonions}
\label{app:quickspec}

As reported in the main text, QuickSpec generated close to 10,000 lemmas on the Octonions example, so we do not include the full output here. However, we include the output on a small subset of the Octonions theory to illustrate some of the problems.

QuickSpec takes as input a list of function and constant symbols out of which it will build the terms. We started by giving it the functions (with $\times$, inverse and $1$). As the inverse of 0 is not defined, we had to instruct QuickSpec to only test using non-zero Octonions, a limitation not shared by $\tool$. For the same reason we could not give it the constant symbol $0$ or the functions $+$ and $-$, which would allow it to construct a zero octonion. The output of QuickSpec is shown in Figure \ref{fig:QuickSpec1}.
\begin{figure}[t]
    \centering
    \caption{Output from QuickSpec on Octonions for multiplication and inverse.}
    \label{fig:QuickSpec1}
\begin{framed}
\tiny
\begin{verbatim}
== Functions ==                  
(*) :: It -> It -> It
  1 :: It

== Laws ==
  1. x * 1 = x                   
  2. 1 * x = x                   
  3. (x * x) * y = x * (x * y)   
  4. (x * y) * x = x * (y * x)
  5. (x * y) * y = x * (y * y)
  6. x * (y * (x * y)) = (x * y) * (x * y)
  7. x * (y * (y * x)) = (x * y) * (y * x)
  8. x * (y * (y * y)) = (x * y) * (y * y)
  9. x * ((y * z) * x) = (x * y) * (z * x)
 10. (x * (y * x)) * z = x * (y * (x * z))
 11. ((x * y) * z) * y = x * (y * (z * y))
                                 
== Functions ==
inv :: It -> It

== Laws ==
 12. inv 1 = 1                   
 13. inv (inv x) = x             
 14. x * inv x = 1               
 15. inv x * inv y = inv (y * x)  
 16. inv x * (x * y) = y           
 17. x * (y * inv x) = (x * y) * inv x
 18. (inv x * y) * x = inv x * (y * x)
\end{verbatim}
\end{framed}
\end{figure}
So far the number of lemmas is manageable and 9 of the 18 are found in the AFP theory, giving a precision of 50\%.

Then we ran QuickSpec again, adding the extra functions $+$, $0$, inner product ($\cdot$) and norm, but removing the inverse operation (as discussed above). The results are still reasonable but, especially with the inner product function, most of the lemmas are uninteresting, not found in the AFP theory, and generated only because they happen to be true (see Figures \ref{fig:QuickSpec2} and \ref{fig:QuickSpec3}).
\begin{figure}[htbp]
    \centering
    \caption{Output from QuickSpec on Octonions with $+$, $0$, inner product ($\cdot$) }
    \label{fig:QuickSpec2}
\begin{framed}
\tiny
\begin{verbatim}
== Functions ==                  
(*) :: It -> It -> It
  0 :: It
  1 :: It

== Laws ==
  1. x * 0 = 0                   
  2. x * 1 = x                   
  3. 0 * x = 0                   
  4. 1 * x = x                   
  5. (x * x) * y = x * (x * y)   
  6. (x * y) * x = x * (y * x)
  7. (x * y) * y = x * (y * y)
  8. x * (y * (x * y)) = (x * y) * (x * y)
  9. x * (y * (y * x)) = (x * y) * (y * x)
 10. x * (y * (y * y)) = (x * y) * (y * y)
 11. x * ((y * z) * x) = (x * y) * (z * x)
 12. (x * (y * x)) * z = x * (y * (x * z))
 13. ((x * y) * z) * y = x * (y * (z * y))
                                   
== Functions ==
(+) :: It -> It -> It

== Laws ==
 14. x + y = y + x                
 15. x + 0 = x                    
 16. (x + x) * y = x * (y + y)     
 17. (x + y) + z = x + (y + z)    
 18. x * (y + 1) = x + (x * y)    
 19. (x + 1) * y = y + (x * y)     
 20. (x + x) * (x * y) = (x * x) * (y + y)
 21. (x + x) * (y * x) = (x * y) * (x + x)
 22. (x + x) * (y * y) = (x * y) * (y + y)
 23. (x * y) + (x * z) = x * (y + z)
 24. (x * y) + (z * y) = (x + z) * y
 25. (x + 1) * (x + x) = (x + x) * (x + 1)
 26. x * (y * (x + y)) = (x * y) * (x + y)
 27. x * (y + (y * x)) = (x * y) * (x + 1)
 28. x * (y + (y * y)) = (x * y) * (y + 1)
 29. (x * (x + y)) * y = x * ((x + y) * y)
 30. ((x + y) * x) * y = (x + y) * (x * y)
 31. (x + (x * x)) * y = (x + 1) * (x * y)
 32. (x + (y * x)) * y = (y + 1) * (x * y)
 33. (x + (x + x)) * y = x * (y + (y + y))
                                     
== Functions ==
(·) :: It -> It -> It

== Laws ==
 34. x · y = y · x                
 35. x · 0 = 0                     
 36. 1 · 1 = 1                     
 37. (x · y) * z = z * (x · y)     
 38. x · (y * x) = x · (x * y)      
 39. x · (y + y) = y · (x + x)      
 40. x · (y · y) = y · (x * y)      
 41. x · (y · 1) = y · (x · 1)      
 42. 1 · (x * y) = 1 · (y * x)      
 43. 1 · (x · y) = x · y            
 44. 1 · (x + 1) = 1 + (x · 1)      
 45. (x · y) + (x · z) = x · (y + z) 
 46. (x * y) · (x * z) = (x · x) * (y · z)
 47. (x * y) · (z * y) = (x · z) * (y · y)
 48. (x * y) · (y + x) = (y * x) · (y + x)
 49. (x * y) · (z · w) = (y * x) · (z · w)
 50. (x · y) · (z · w) = (x · y) * (z · w)
 51. (x + x) * (x · 1) = (x * x) + (x · x)
 52. (x · y) * (z · 1) = z · (x · y)
 53. (x * y) · (y + 1) = (y * x) · (y + 1)
 54. (x · y) · (z · 1) = z · (x · y)
 55. (1 + 1) · (1 + 1) = (1 + 1) * (1 + 1)
 56. x * (y * (z · w)) = (x * y) * (z · w)
 57. (x * (y · z)) * w = (x * w) * (y · z)
 58. (x + (y · z)) * x = x * (x + (y · z))
 59. x · (y * (z * x)) = (y * z) · (x · x)
 60. x · (y * (z · w)) = (x · y) * (z · w)
 61. x · ((x * y) * z) = (y * z) · (x · x)
 62. x · ((y * x) * z) = x · (y * (x * z))
 63. x · ((x + y) * y) = x · (y * (x + y))
 64. x · (y + (z * x)) = x · (y + (x * z))
 65. x · (y + (x · x)) = x · (y + (x * x))
 66. x · (y + (x · y)) = (x + 1) · (x · y)
 67. x · (y + (y · y)) = y · (x + (x * y))
 68. x · (y · (z * y)) = z · (y · (x * y))
 69. x · (y · (z · w)) = y · (x · (z · w))
 70. x · (y + (y + y)) = y · (x + (x + x))
 71. x · (y * (z · 1)) = z · (x · y) 
 72. 1 · ((x * y) * z) = 1 · (x * (y * z))
 73. (x * (y · 1)) · z = y · (x · z) 
 74. x · (y + (y · 1)) = y · (x + (x · 1))
\end{verbatim}
\end{framed}
\end{figure}
\begin{figure}[tbp]
    \centering
    \caption{Continued output from Figure \ref{fig:QuickSpec2}, Octonions with $+$, $0$, inner product ($\cdot$) and norm}
    \label{fig:QuickSpec3}
\begin{framed}
\tiny
\begin{verbatim}
== Functions ==
norm :: It -> It

== Laws ==
 75. norm 0 = 0                   
 76. norm 1 = 1                   
 77. norm x = x · x                
 78. norm (x + (x * x)) = norm (x + norm x)
 79. (x · y) * (z · 1) = z · (x · y) 
 80. (x · y) · (z · 1) = z · (x · y) 
 81. norm (norm x + (y * z)) = norm (norm x + (z * y))
 82. (norm x + norm y) * z = z * (norm x + norm y)
 83. x · (y * (z · 1)) = z · (x · y)  
 84. (x * (y · 1)) · z = y · (x · z)  
 85. norm x * (1 + norm y) = norm x + norm (x * y)
 86. norm x · (1 + norm y) = norm x + norm (x * y)
 87. norm (x * (x + norm x)) = norm (norm x * (x + 1))
\end{verbatim}
\end{framed}
\end{figure}
In general, it seems that the more function symbols there are, the more QuickSpec suffers from combinatorial blowup and generating uninteresting lemmas.

In the complex numbers, the functions $\operatorname{Re}, \operatorname{Im} : \mathbb{C} \rightarrow \mathbb{R}$ extract the real and imaginary parts of a number respectively. In the octonions, there are eight analogous functions $\operatorname{Re}, \operatorname{Im_1}\ldots\operatorname{Im_7} : \mathbb{O} \rightarrow \mathbb{R}$. When we added these functions, QuickSpec generated 775 laws, the vast majority extremely uninteresting for a human. An example of a typical generated law is $\operatorname{Re} (x + \operatorname{Im_2}\, y \times z) = \operatorname{Re} (x + y \times \operatorname{Im_2}\, z)$.

Corresponding to the complex number $i$ there are seven octonions $e_1 \ldots e_7$ which (together with $1$) act as unit numbers. Unfortunately when we add these as constants then QuickSpec starts to generate many thousands of irrelevant lemmas. Examples include: 
$$(e_5 - e_4) \times (e_1 + e_4) = (1 + e_1) \times (e_4 - e_1)$$ and 
$$e_4 \times (\operatorname{Im_1}\, x \times \operatorname{Im_2}\, y) = \operatorname{Im_7} (\operatorname{Im_1}\, x \times (y \times e_4))$$ The problem is that QuickSpec does not attempt to judge which lemmas are relevant to a human user.